\documentclass[11pt]{article}

\usepackage{acl}

\usepackage{times}
\usepackage{latexsym}
\usepackage[T1]{fontenc}
\usepackage[utf8]{inputenc}
\usepackage{microtype}
\usepackage{inconsolata}
\usepackage{graphicx}
\usepackage{amsmath}
\usepackage{amssymb}
\usepackage{booktabs}
\usepackage{multirow}
\usepackage{colortbl}
\usepackage{threeparttable}
\usepackage{algorithm}
\usepackage{algpseudocode}
\usepackage{pifont}

\definecolor{transgray}{gray}{0.90}
\graphicspath{{figures/}}

\title{Speculate While You Reason: Teaching Agents to Predict Their Next Tool Call via Joint Agent--Speculator RL}

\author{
 Jiabao Ji$^1$\thanks{Equal contribution. Work done when Jiabao, Yujian, Li, were interning at Linkedin Inc.}
 \quad 
    Yujian Liu$^{1*}$ \quad
    Li An $^{1*}$
    \\
    \textbf{Rohit Jain\textsuperscript{2}}\quad 
    \textbf{Gungor Polatkan\textsuperscript{2}}\quad 
    \textbf{Siyu Zhu\textsuperscript{2}} \quad
    \textbf{Shiyu Chang\textsuperscript{1}}\\
    \textsuperscript{1}University of California, Santa Barbara \quad 
    \textsuperscript{2}Linkedin Inc \\
\texttt{\{jiabaoji,yujianliu,li\_an,chang87\}@ucsb.edu}
}

\let\cite\citep

\begin{document}
\maketitle

\begin{abstract}
    Large language model agents often spend substantial wall-clock time waiting for tool call results. Tool-call speculation can hide this latency by predicting and pre-executing an agent's next tool call if the prediction matches the agent's eventual tool call, but existing speculators are typically separate draft models or cached traces that are poorly aligned with the deployed agent's own behavior. We identify this \emph{speculator--agent gap} and show that the target agent itself is a strong next-call speculator.
    This points to a simpler design: unifying the agent and speculator within the same model. In this paper, we introduce the \emph{self-speculating agent}, a single model that both solves tasks in agent mode and predicts its next tool call from partial trajectories in speculator mode, fully reusing prefix KV cache. To enable this dual-mode agent without degrading performance, we propose a joint agent-speculator reinforcement learning method, which derives speculation targets from the agent's own rollouts and alternates agent and speculator updates. Across agentic search QA and conversational tool-use agentic tasks, our method improves average next tool-call Hit@1 from 44.1 to 61.2 for Qwen3-4B and from 48.9 to 66.3 for Qwen3.5-4B, while preserving agent task success.
\end{abstract}

\section{Introduction}
\label{sec:intro}

\begin{figure}[t]
    \centering
    \includegraphics[width=0.95\linewidth]{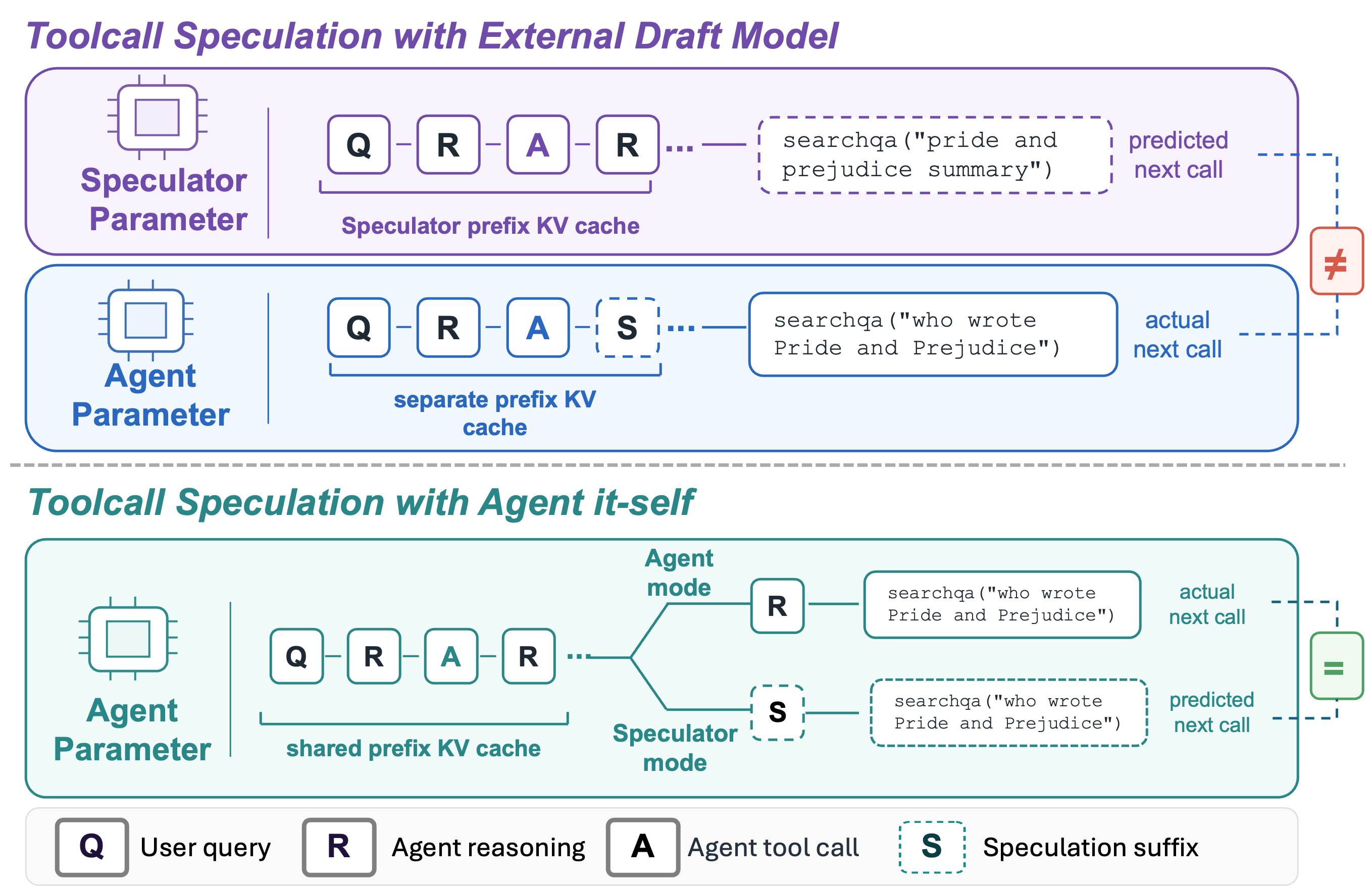}
    \vspace{1mm}
    \caption{External versus self tool-call speculation. The upper panel shows
    an external draft model with separate parameters and KV cache, which 
    predict a different next call due to the speculator--agent gap. The lower panel shows
    self-speculation with a shared prefix KV cache.}
    \label{fig:self-speculating-overview}
    \vspace*{-0.2in}
\end{figure}

Large language model (LLM) agents solve complex tasks by interleaving
natural-language reasoning with calls to external tools, including web search
engines, databases, code interpreters, and task-specific APIs~\citep{schick2023toolformer,patil2023gorilla,chen2023fireact,liu2024apigen}.
Recent work further trains and evaluates such agents in search, web,
function-calling, and long-horizon interaction settings~\citep{jin2025searchr1,song2025r1searcher,tan2025ragr1,qi2025webrl,qian2025toolrl,zhang2025toolr1,liu2026agentskills,luo2025agentlightning}.
These tools extend agents beyond their parametric knowledge by allowing them to
retrieve up-to-date information, execute computations, inspect structured
records, and interact with external environments. This flexibility, however,
comes at the cost of latency. Unlike token generation, tool calls often involve
remote services, network I/O, or even invoking another LLM as a sub-agent. As a result, agents may spend a substantial fraction of
their inference time waiting for tool results rather than performing token generation~\citep{nichols2025speculativetoolcalls,huang2025spagent,hooper2026speculativeinteraction},
making reduced waiting time central to responsive agents.

A promising approach is \emph{tool-call speculation}: issue likely future tool
calls before the agent explicitly produces them. A speculator observes an
intermediate trajectory, predicts the agent's next tool call, and executes that
call in parallel while the agent continues reasoning. If the prediction exactly
matches the agent's eventual tool call, the cached tool call result can be reused,
hiding the tool call latency. If it does not match, the
pre-executed result must be discarded. This exact-match reuse scenerio makes
tool-call speculation different from general tool use. Instead of predicting a general tool call, 
the speculator must produce the particular tool call that the deployed agent will actually take.

This requirement makes the design of the speculator especially important. Prior
work typically builds the speculator outside the deployed agent, either by using
a smaller LLM as an external draft model~\citep{ye2026speculativeactions,nichols2025speculativetoolcalls,huang2025spagent}
or by consulting cached tool call traces derived from
previous agent trajectories~\citep{zhong2026dualspec,sui2026paste}. These methods
show promising results to reduce latency, but they share a
fundamental limitation: the speculator is not the agent. An off-the-shelf draft
model approximates a generic tool-using assistant rather than the particular
agent being deployed. A cache-based method replays calls from previous
trajectories, which may correspond to different user queries, different
intermediate reasoning states, or outdated environments. In both cases, the
speculation methods remain separated from the actual deployed agent.

We refer to this separation as the \emph{speculator--agent gap}: conditioned on
the same intermediate agent trajectory, the speculator and the deployed agent may
choose different next tool calls. Figure~\ref{fig:self-speculating-overview}
illustrates this gap, where an external draft model maintains a separate set of model parameters and KV-cache
and predicts a different call than the target agent eventually issues. In
Section~\ref{sec:empirical-self-speculate}, we find that the gap appears even
within the same model family: smaller Qwen draft models often fail to match the
4B target agent's next tool call, while the 4B agent itself is already a stronger off-the-shelf speculator when prompted to predict
its own next call. 
External speculators also introduce practical systems overhead, including an additional
model weights, a separate KV cache, or historical traces.

These observations suggest a simpler speculator design: rather than aligning a separate
speculator to the agent, the agent itself can speculate. This leads to the
central research question: \emph{Can we train a agent to better
anticipate its own next tool call while preserving task solving performance?}

We therefore present a \emph{self-speculating agent}: a single model trained to
operate in two modes. In \emph{agent mode}, the model follows the standard
agentic trajectory, producing reasoning, tool calls, observations, and final
answers. In \emph{speculator mode}, the same model receives an intermediate
trajectory together with a short speculation suffix and directly predicts the
next structured tool call. Because both modes share the trajectory prefix,
inference can reuse the agent's prefix KV cache and branch only for the
speculation suffix and candidate call. Unlike external speculation, this
approach does not require serving an auxiliary draft model or maintaining a
historical tool-call traces.

However, training such an agent is nontrivial. Self-speculation couples two
objectives that are naturally imbalanced: the agent must remain effective at
solving tasks, while also becoming accurate at predicting its own next tool call
from partial trajectories. The two modes differ substantially in output
distribution, and naively training the agent to improve speculation can hurt
agent performance. To address this problem, we introduce \emph{joint agent-speculator RL}
During the RL process, we sample agent rollouts from the current policy, construct
speculator queries from those rollouts, and optimize each mode in an alternating procedure.
This keeps the speculator mode on-policy: it is trained on the same contexts that arise from the current agent, 
rather than on stale rollouts or data from a different model. We further stabilize this joint training with several training-time techniques, such as optimizer resets at mode switches.

Experiments across agentic search QA and conversational tool
use agentic tasks show that joint agent-speculator RL substantially improves speculation performance
while preserving task performance. For Qwen3-4B and Qwen3.5-4B, average 
next tool-call Hit@1 score improves from 44.1 to 61.2 and from 48.9 to 66.3, respectively,
while average downstream task success remains stable or slightly improves.

Our contributions are as follows:
\begin{itemize}
    \item We identify the speculator--agent gap in agentic tool-use tasks where off-the-shelf draft models often fail to match the agent's next structured call and add serving overhead.
    \item We introduce a self-speculating agent, where one LLM switches between task-solving and next-call prediction modes.
    \item We propose joint agent-speculator RL, which trains speculation from the current agent's own rollouts while preserving task success.
\end{itemize}

\section{Evaluating Off-the-Shelf Models for Tool-Call Speculation}
\label{sec:empirical-self-speculate}

Off-the-shelf draft models are a natural starting point for tool-call speculation. We
evaluate their speculator--agent gap in realistic agentic settings, where the
speculator observes the deployed agent's context immediately before a tool call
and predicts the next tool call, including both the tool name and
arguments. This setup measures whether an off-the-shelf model can match the
deployed agent's tool decisions under the same inference-time context.

\subsection{Evaluation Protocol}
\label{sec:empirical-self-speculate:protocol}

\begin{figure}[t]
\centering
\includegraphics[width=0.92\linewidth]{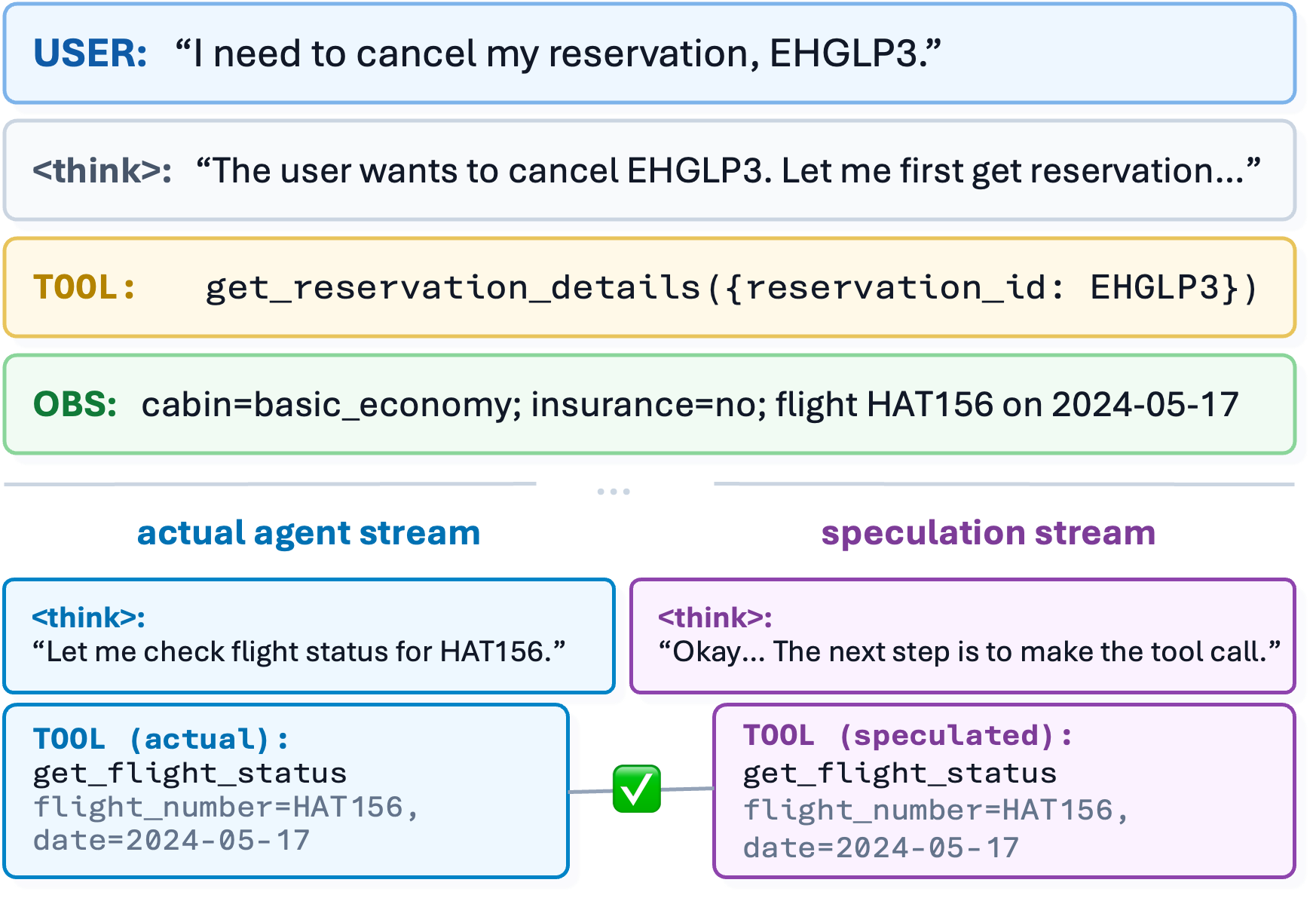}
\vspace{1mm}
\caption{Online tool-call speculation example. From the shared trajectory
prefix, the left stream continues to the agent's reasoning and actual tool call generation, while the
right stream predicts the tool call in parallel. 
}
\label{fig:speculation-rollout-example}
\vspace*{-0.12in}
\end{figure}

For each target agent, we run speculation online during the agent's rollout. 
In each domain, we sample 50 evaluation queries and execute the deployed agent in the environment. 
Whenever the rollout reaches a tool-call boundary, the serving controller switches from the target
agent model to an off-the-shelf speculator model. 
The speculator receives the current trajectory prefix with a fixed speculation suffix appended:
\begin{quote}
\small\ttfamily
\textless think\textgreater{} Okay, let's see. The user provided what I need.
I'll look it up. The next step is to make the tool call.
\textless/think\textgreater{}
\end{quote}
The speculator then decodes one candidate tool call. The controller switches
back to the target agent, lets the agent produce the actual call, and executes
the agent's call in the environment. Thus speculation is evaluated under the
same online state that the agent sees, while the environment state is still
determined only by the deployed agent.
Figure~\ref{fig:speculation-rollout-example} illustrates this online comparison in a
$\tau$-bench trajectory: the speculator is prompted from the current prefix, the
agent continues independently, and the pre-executed result is reusable only when
the speculated tool name and arguments exactly match the agent's eventual call.

We instantiate this protocol in two same-family comparisons. For Qwen3-4B agent, we
evaluate Qwen3-0.6B and Qwen3-1.7B as speculators; for Qwen3.5-4B agent, we
evaluate Qwen3.5-0.8B and Qwen3.5-2B. In both families, we also prompt the
target 4B model to predict its own next call from the same online prefix. 
We run the protocol on 50 queries from MuSiQue in the agentic SearchQA environment~\citep{jin2025searchr1} and airline domain of $\tau$-bench~\citep{yao2024taubench,barres2025tau2bench}.

Table~\ref{tab:offshelf-results} reports both prediction quality and serving
cost. We use Hit@1 exact match as the primary speculation metric because a
pre-executed tool call is reusable only when both the tool name and argument
dictionary match the agent's eventual call. We also report speculation
wall-clock time and peak GPU memory to capture the overhead of alternating
between the target agent and a separate draft model during the same rollout. The
memory estimate includes both resident model weights and peak KV-cache state:
external speculators require the target agent weights, the draft model weights,
and two active KV caches, whereas self-speculation reuses the target model
weights and the target prefix KV cache, adding only the small speculation branch.
All measurements in this off-the-shelf setting use a single H100 GPU. We
implement the comparison in the SGLang inference engine with an explicit
parameter switch between the external drafter model and the target agent model,
so each query serially alternates agent generation, speculator generation, and
agent continuation while recording time and memory usage.

\begin{table}[t]
\centering
\renewcommand{\arraystretch}{1.1}
\small
\setlength{\tabcolsep}{2.5pt}
\resizebox{0.95\columnwidth}{!}{%
\begin{tabular}{@{}l|ccc|ccc@{}}
\toprule \midrule
\multirow{2}{*}{Speculator} &
\multicolumn{3}{c|}{\textbf{MuSiQue}} &
\multicolumn{3}{c}{\textbf{$\tau$-bench}} \\
\cmidrule(lr){2-4} \cmidrule(lr){5-7}
 & H@1$\uparrow$ & Time$\downarrow$ & Mem.$\downarrow$
 & H@1$\uparrow$ & Time$\downarrow$ & Mem.$\downarrow$ \\
\midrule
\rowcolor{transgray}
\multicolumn{7}{c}{\textbf{Target agent: Qwen3-4B}} \\
\midrule
Qwen3-0.6B & 4.3 & 27.4 & 10.71 & 16.8 & 45.2 & 14.59 \\
Qwen3-1.7B & 14.7 & 33.1 & 12.76 & 25.4 & 52.1 & 16.64 \\
Qwen3-4B self & \textbf{25.3} & \textbf{8.3} & \textbf{8.70} & \textbf{37.2} & \textbf{24.1} & \textbf{10.90} \\
\midrule
\rowcolor{transgray}
\multicolumn{7}{c}{\textbf{Target agent: Qwen3.5-4B}} \\
\midrule
Qwen3.5-0.8B & 8.7 & 22.3 & 9.31 & 21.3 & 39.2 & 9.98 \\
Qwen3.5-2B & 16.2 & 30.6 & 11.55 & 32.8 & 54.1 & 12.22 \\
Qwen3.5-4B self & \textbf{27.4} & \textbf{6.7} & \textbf{7.73} & \textbf{44.1} & \textbf{19.1} & \textbf{8.22} \\
\midrule
\bottomrule
\end{tabular}
}
\caption{Online off-the-shelf speculation results. For each target 4B agent, we
compare smaller same-family external draft speculators with the target model
prompted to speculate its own next tool call. GPU memory includes model weights
and peak KV-cache.}
\label{tab:offshelf-results}
\vspace*{-0.12in}
\end{table}

\subsection{Findings}
\label{sec:empirical-self-speculate:findings}

\paragraph{External speculators add serving overhead.} Even small draft models
require additional weights and a second KV cache, so their cost is not just the
extra decoding step. In Table~\ref{tab:offshelf-results}, Qwen3-4B on MuSiQue
increases from 8.70GB and 8.3s with self-speculation to 10.71GB/27.4s with
Qwen3-0.6B and 12.76GB/33.1s with Qwen3-1.7B. The same pattern appears in
$\tau$-bench and in the Qwen3.5 series. External speculation therefore needs
large Hit@1 gains to justify the extra serving cost, especially when memory is
limited or agentic contexts are long.

\paragraph{The agent is the strongest off-the-shelf speculator for itself.} The
target agent outperforms smaller same-family draft models, indicating that
family similarity alone does not recover the target agent's next-call
distribution. For Qwen3-4B, self-speculation reaches 25.3 Hit@1 on MuSiQue and
37.2 on $\tau$-bench, compared with 14.7 and 25.4 for the strongest external
speculator. For Qwen3.5-4B, self-speculation improves over Qwen3.5-2B from 16.2
to 27.4 on MuSiQue and from 32.8 to 44.1 on $\tau$-bench. This supports the
speculator--agent gap: the best off-the-shelf predictor of an agent's next call
is often the agent itself. These findings motivate training the agent itself as
the speculator instead of deploying a separate external LLM.

\section{Method}
\label{sec:method}

Motivated by the findings in
Section~\ref{sec:empirical-self-speculate:findings}, we train a single
tool-calling LLM agent $\pi_\theta$ to act both as the task-solving agent and as
its own next-call speculator. We first define the
agent task reward and the speculation reward used to score predicted tool calls.
Second, we describe the joint agent--speculator RL method, which turns
current agent trajectories into on-policy speculation targets. Finally, we describe 
the training recipe used to stabilize the shared policy for both modes.

\subsection{Task and Reward Formulation}
\label{sec:method:task-formulation}

Given a user query $x$, the agent interacts with an environment through a
sequence of reasoning steps, tool calls, and tool observations. We denote an
agent trajectory by
$\tau=(x,r_1,a_1,o_1,\ldots,r_T,a_T,o_T,y)$, where $r_t$ is the reasoning
before the $t$-th tool call, $a_t$ is the tool call, $o_t$ is the returned
observation, and $y$ is the final answer. Each tool call is a structured action
$a_t=(n_t,\sigma_t)$, with tool name $n_t$ and argument dictionary $\sigma_t$.
The agent reward $R_{\mathrm{ag}}(\tau)$ is evaluated on the final answer
$y$ against the ground-truth answer $y^*$ defined in the environment.

The speculation task is derived from the same trajectory and is implemented by
the same policy $\pi_\theta$. Agent and speculator modes therefore share model
parameters, with the speculator mode triggered simply by appending a fixed speculation suffix 
to the input context. For each tool-call turn $t$, we take the prefix immediately before
the call,
$h_t=(x,r_1,a_1,o_1,\ldots,a_{t-1})$, append a fixed speculation suffix
$s_{\mathrm{sp}}$, and ask the policy to emit the next structured tool
call, $\hat a_t\sim\pi_\theta(\cdot\mid h_t\oplus s_{\mathrm{sp}})$. The target
is the agent's own next call $a_t$ from the sampled trajectory. Since $a_t$
comes from the current policy, the targets change as the agent changes, keeping
speculation training on-policy with the deployed agent.

We score a predicted call $\hat a_t=(\hat n_t,\hat\sigma_t)$ against the target
call $a_t=(n_t,\sigma_t)$ using a name-conditioned argument reward. The tool-name
score is $R_{\mathrm{name}}=\mathbf{1}[\hat n_t=n_t]$. If the tool name matches,
we compare argument dictionaries with macro token-F1 over target argument keys:
\[
R_{\mathrm{args}}
= \frac{1}{|\mathrm{keys}(\sigma_t)|}
\sum_{k\in\mathrm{keys}(\sigma_t)}
\mathrm{F1}\!\left(\hat\sigma_t[k],\sigma_t[k]\right).
\]
Missing predicted arguments receive score zero, and extra predicted arguments
do not receive credit. Argument values are normalized to strings before
token-F1. For tools with no arguments, we set $R_{\mathrm{args}}=1$ after a
correct tool-name match. The final speculation reward is
\begin{equation}
R_{\mathrm{sp}} = R_{\mathrm{name}}\cdot R_{\mathrm{args}}.
\label{eq:reward}
\end{equation}
This construction matches the exact-match reuse requirement for speculation: a wrong tool name cannot
reuse the pre-executed result, while a correct tool name can still receive
partial credit for close argument values. Macro averaging gives each target
argument key equal weight, avoiding pooled token-F1 behavior where long argument
values dominate shorter but equally important keys.

\subsection{Joint Agent--Speculator RL}
\label{sec:method:dual-mode-rollout}

Agent and speculator samples differ substantially. The agent mode generates long,
reasoning-heavy trajectories, whereas the speculator mode generates a single structured tool call 
from a partial trajectory. Their different output distributions and rewards make direct joint training difficult.
We therefore keep the data collection coupled but the policy updates separated. 
Each iteration constructs fresh on-policy data for both modes, while each optimizer step uses only the objective selected for
the current training iteration.

At RL iteration $k$, we sample a query batch $Q\subset\mathcal{D}$ and generate
$G_{\mathrm{ag}}$ agent trajectories for each $x\in Q$. These trajectories serve as standard agent rollouts: their final rewards
$R_{\mathrm{ag}}$ are normalized within each query group to produce
DAPO-style relative advantages for the agent objective~\citep{yu2025dapo}. At
the same time, they provide the speculation supervision induced by the current
policy. For each realized tool-call turn, we take the prefix $h_t$ and the
actual next call $a_t$ from the sampled trajectory, form the speculator input
$h_t\oplus s_{\mathrm{sp}}$, and treat $a_t$ as the target call.

A fixed schedule $m(k)\in\{\mathrm{ag},\mathrm{sp}\}$ selects which objective is
optimized at iteration $k$. If $m(k)=\mathrm{ag}$, we update
$\pi_\theta$ with the agent rollouts and advantages from $R_{\mathrm{ag}}$.
If $m(k)=\mathrm{sp}$, we update the same $\pi_\theta$ with the speculator
rollouts and advantages from $R_{\mathrm{sp}}$. Both modes use the same
DAPO policy optimization~\citep{yu2025dapo}; they differ in the input context, rollout structure,
and reward function. In our final runs, $m(k)$ follows a repeated 4:8
pattern: four consecutive agent updates followed by eight consecutive
speculator updates. Algorithm~\ref{alg:training-recipe} summarizes the full procedure.

\subsection{Stabilizing Dual-Mode RL}
\label{sec:method:training-recipe}

Training a shared policy for both task solving and self-speculation is
non-trivial. The two modes produce different output distributions and optimize
different rewards, so direct joint training can suffer from mode interference or
collapse into unproductive tool-calling behavior. We therefore use a training
recipe with three components. \ding{182} Before RL, we run a short SFT warmup
on a mixture of successful agent trajectories and their corresponding
intermediate-trajectory speculation examples. The agent examples supervise the
full successful trajectory, while each speculation example takes a partial
trajectory, appends $s_{\mathrm{sp}}$, and supervises the model to produce the
ground-truth next tool call. This gives the model well-formed behavior in both
modes before RL begins. \ding{183} We reset optimizer state whenever the
schedule switches between agent and speculator updates. Since the two modes optimize different rewards, 
momentum and adaptive statistics differ between the two modes, which can strongly affect subsequent optimization
steps~\citep{sutskever2013momentum,kingma2015adam}; the reset reduces carryover
from one mode's objective to the next. \ding{184} The agent reward includes a
penalty for repeated meaningless tool calls, discouraging a common failure mode
in which the agent continues calling tools without making task progress.

\begin{algorithm}[t]
\small
\begin{algorithmic}[1]
\Require Policy $\pi_\theta$, query set $\mathcal{D}$, speculation suffix $s_{\mathrm{sp}}$, alternation schedule $m(k)$, group sizes $G_{\mathrm{ag}},G_{\mathrm{sp}}$
\State $m_{\mathrm{prev}}\gets \varnothing$
\For{RL iteration $k = 1,2,\ldots$}
    \State Sample query batch $Q \sim \mathcal{D}$
    \State Generate agent groups $\mathcal{T}=\{\tau_i(x)\}_{x\in Q,\,i=1}^{G_{\mathrm{ag}}}$
    \State Build speculation pairs $\mathcal{S}=\{(h_t,a_t)\!: t\in\tau,\tau\in\mathcal{T}\}$
    \State Generate speculator groups $\widehat{\mathcal{A}}=\{\hat a_{t,j}\sim\pi_\theta(\cdot\mid h_t\oplus s_{\mathrm{sp}})\}_{(h_t,a_t)\in\mathcal{S},\,j=1}^{G_{\mathrm{sp}}}$
    \If{$m(k)\ne m_{\mathrm{prev}}$}
        \State Reset optimizer state
    \EndIf
    \If{$m(k)=\mathrm{ag}$}
        \State DAPO update $\pi_\theta$ on grouped agent rewards $R_{\mathrm{ag}}(\mathcal{T})$
    \Else
        \State DAPO update $\pi_\theta$ on grouped speculation rewards $\{R_{\mathrm{sp}}(\hat a_{t,j},a_t)\}_{(h_t,a_t)\in\mathcal{S},\,j=1}^{G_{\mathrm{sp}}}$
    \EndIf
    \State $m_{\mathrm{prev}}\gets m(k)$
\EndFor
\end{algorithmic}
\caption{Joint Agent-Speculator RL}
\label{alg:training-recipe}
\end{algorithm}

\begin{table*}[!t]
\centering
\renewcommand{\arraystretch}{1.12}
\small
\setlength{\tabcolsep}{1.2pt}
\resizebox{0.95\textwidth}{!}{%
\begin{tabular}{@{}l|cc|cc|cc|cc|cc|cc@{}}
\toprule \midrule
\multirow{2}{*}{Model} &
\multicolumn{2}{c|}{\textbf{HotpotQA}} &
\multicolumn{2}{c|}{\textbf{MuSiQue}} &
\multicolumn{2}{c|}{\textbf{BCP}} &
\multicolumn{2}{c|}{\textbf{$\tau$-bench Airline}} &
\multicolumn{2}{c|}{\textbf{$\tau$-bench Retail}} &
\multicolumn{2}{c}{\textbf{Avg.}} \\
\cmidrule(lr){2-3} \cmidrule(lr){4-5} \cmidrule(lr){6-7}
\cmidrule(lr){8-9} \cmidrule(lr){10-11} \cmidrule(lr){12-13}
 & H@1$\uparrow$ & Succ.$\uparrow$
 & H@1$\uparrow$ & Succ.$\uparrow$
 & H@1$\uparrow$ & Succ.$\uparrow$
 & H@1$\uparrow$ & Succ.$\uparrow$
 & H@1$\uparrow$ & Succ.$\uparrow$
 & H@1$\uparrow$ & Succ.$\uparrow$ \\
\midrule
\rowcolor{transgray}
\multicolumn{13}{c}{\textbf{Qwen3 series}} \\
\midrule
Qwen3-0.6B & 3.6 & 52.1 & 4.2 & 15.1 & 2.8 & 7.0 & 16.5 & 21.6 & 18.5 & 36.4 & 9.1 & 26.4 \\
Qwen3-1.7B & 13.7 & 52.1 & 15.8 & 15.1 & 11.9 & 7.0 & 24.1 & 21.6 & 27.3 & 36.4 & 18.6 & 26.4 \\
Qwen3-4B & 23.2 & 52.1 & 26.4 & 15.1 & 20.7 & 7.0 & 35.6 & 21.6 & 39.7 & 36.4 & 29.1 & 26.4 \\
Qwen3-4B-SFT & 37.8 & 51.3 & 40.9 & 16.4 & 35.6 & 7.4 & 51.2 & 22.0 & 55.1 & 35.8 & 44.1 & 26.6 \\
Qwen3-4B-RL & \textbf{55.2} & \textbf{53.0} & \textbf{58.4} & \textbf{17.1} & \textbf{52.1} & \textbf{8.1} & \textbf{68.0} & \textbf{23.1} & \textbf{72.3} & \textbf{37.0} & \textbf{61.2} & \textbf{27.7} \\
\midrule
\rowcolor{transgray}
\multicolumn{13}{c}{\textbf{Qwen3.5 series}} \\
\midrule
Qwen3.5-0.8B & 8.4 & 63.1 & 9.5 & 25.1 & 7.8 & 29.6 & 20.2 & 56.4 & 22.4 & 69.7 & 13.7 & 48.8 \\
Qwen3.5-2B & 14.9 & 63.1 & 16.8 & 25.1 & 13.7 & 30.2 & 31.6 & 56.4 & 34.7 & 69.7 & 22.3 & 48.9 \\
Qwen3.5-4B & 25.4 & 63.1 & 28.3 & 25.1 & 23.6 & 30.7 & 42.1 & 56.4 & 46.8 & 69.7 & 33.2 & 49.0 \\
Qwen3.5-4B-SFT & 40.7 & 62.2 & 43.9 & 26.0 & 38.2 & 31.2 & 58.6 & 56.4 & 62.9 & 70.2 & 48.9 & 49.2 \\
Qwen3.5-4B-RL & \textbf{58.6} & \textbf{64.3} & \textbf{61.8} & \textbf{27.0} & \textbf{55.7} & \textbf{32.4} & \textbf{76.1} & \textbf{58.1} & \textbf{79.4} & \textbf{71.2} & \textbf{66.3} & \textbf{50.6} \\
\midrule
\bottomrule
\end{tabular}
}
\caption{Main self-speculation results. We compare smaller same-family
speculators, the base 4B agent, the SFT-initialized 4B agent, and our dual-mode
RL agent across search and $\tau$-bench benchmarks. For external speculator
rows, task success is the success of the fixed target 4B agent whose next calls
are being predicted, not the smaller speculator run as an independent agent.}
\label{tab:main-self-speculation}
\vspace{-0.2in}
\end{table*}

\section{Experiments}
\label{sec:setup}

Our experiments evaluate whether a single model can improve its own
next-tool-call speculation while preserving downstream task performance. We
first describe the evaluation setup, then report main results across search and
conversational tool-use benchmarks, analyze cross-domain generalization, and
ablate the stabilization techniques used in joint agent-speculator RL.

\subsection{Experimental Setup}
\label{sec:experiments:setup}

\paragraph{Tasks and environments.}
We evaluate on two families of structured tool-use tasks. Agentic
SearchQA~\citep{jin2025searchr1,song2025r1searcher} requires the model to
answer multi-hop questions by issuing search calls and reasoning over returned
evidence. $\tau$-bench conversational tool-use
tasks~\citep{yao2024taubench,barres2025tau2bench} require the agent to interact
with a simulated user and complete airline or retail tasks through
schema-constrained APIs. These settings stress complementary aspects of
speculation: SearchQA emphasizes query formulation under partial information,
whereas $\tau$-bench emphasizes exact API selection and argument filling in
long-horizon interactions.

\paragraph{Training and evaluation data.}
For Agentic SearchQA settings, we train on FlashQA and evaluate on HotpotQA,
MuSiQue, and BrowseComp-Plus (BCP)~\citep{chen2025browsecompplus}.
For BCP, we follow~\citet{sun2025contextfolding} and use the 150-query evaluation split from their
680/150 train--test partition. We also follow the official BCP setup 
to use Qwen3-Embed-0.6B as the dense retriever. For
conversational tool-use tasks, we train on
ToolScale~\citep{su2025toolorchestra} and evaluate on $\tau$-bench Airline and
$\tau$-bench Retail~\citep{barres2025tau2bench}.

\paragraph{Models and baselines.}
We instantiate our method with Qwen3-4B~\citep{yang2025qwen3} and
Qwen3.5-4B~\citep{qwen2026qwen35modelcard}. We compare against smaller same-family
LLMs as the external speculators, the base 4B agent prompted to predict its own next call,
and the SFT-warmup checkpoint before RL. Specifically, the external speculators are
Qwen3-0.6B and Qwen3-1.7B for Qwen3-4B, and Qwen3.5-0.8B and Qwen3.5-2B for
Qwen3.5-4B. We employ same speculation suffix for all models. 

\paragraph{Training protocol.}
We follow the joint agent-speculator RL recipe in
Section~\ref{sec:method:training-recipe}. Each run begins with a short SFT
warmup on a mix of successful trajectories and speculation examples so that the model can emit well-formed tool
calls under the speculation suffix before RL begins. We then instanstite the alternative RL between
agent updates and speculator updates, resetting optimizer state when the
schedule switches modes. Unless otherwise noted, we use DAPO-style grouped
policy optimization~\citep{yu2025dapo} with $G_{\mathrm{ag}}=8$ agent rollouts,
$G_{\mathrm{sp}}=8$ speculator rollouts, learning rate $3\times10^{-6}$, and a
fixed 200-step budget per RL training. Appendix~\ref{sec:appendix:training-hyperparameters}
lists the remaining implementation details, such as the training hardware.


\paragraph{Evaluation protocol and metrics.}
For speculation evaluation, we first run the model in agent mode and collect
complete trajectories. At each tool-call turn, we truncate the trajectory
immediately before the call, append the speculation suffix, and ask the
speculator to predict the next call. We report Hit@1 exact match, where a hit
requires both the tool name and the full argument dictionary to match the
agent's eventual call. For task quality, we report task success on HotpotQA,
MuSiQue, BCP, $\tau$-bench Airline, and $\tau$-bench Retail. For external
speculators, task success refers to the fixed target 4B agent whose calls are
being predicted, not to the smaller draft model as an independent agent.
Together, these metrics capture the central trade-off: speculation is useful
only if it improves reusable call prediction without degrading task behavior.

\subsection{Main Results}
\label{sec:results}

Table~\ref{tab:main-self-speculation} reports the main results across both model
families and evaluation domains. We highlight the following observations:

\paragraph{Joint agent-speculator RL improves speculation performance.}
Table~\ref{tab:main-self-speculation} extends the finding from
Section~\ref{sec:empirical-self-speculate}: the deployed agent is already a
stronger speculator for itself than smaller same-family draft models. In the
Qwen3 series, the strongest external speculator averages 18.6 Hit@1, while the
base 4B agent reaches 29.1 when prompted to predict its own next call. 
We also highlight that the SFT warmup improves this self-speculation ability, raising average Hit@1 to
44.1 for Qwen3-4B and 48.9 for Qwen3.5-4B. The joint agent-speculator RL method further
improves average Hit@1 to 61.2 and 66.3, respectively, with gains in the same direction on every benchmark. 
Since Hit@1 requires both the tool name and full argument dictionary to match, 
these gains indicate better speculation timing gains.

\paragraph{Agent performance is largely preserved.}
Since the same parameter set is used in both agent and speculator modes for our self-sepculative agent, improving
next-call prediction can risk degrading task solving performance. In
Table~\ref{tab:main-self-speculation}, we observe that task success remains stable after RL,
average success changes from 26.6 to 27.7 for Qwen3-4B and from 49.2 to 50.6
for Qwen3.5-4B. These results confirm the same pattern, with no
systematic degradation across search QA or conversational tool-use benchmarks.
Thus, the speculation gains above are not obtained by sacrificing the deployed
agent's end-task performance; our joint agent-speculator RL achieves accurate
speculation while keeping the agent policy usable.

\subsection{Generalization Across Domains}
\label{sec:analysis:generalization}

An interesting question to ask is whether the ability to self-speculate
generalizes across domains. We investigate this by evaluating checkpoints
trained on one agent task in the other: the SearchQA-trained Qwen3-4B checkpoint
is tested on $\tau$-bench, and the ToolScale-trained Qwen3.5-4B checkpoint is
tested on SearchQA. Table~\ref{tab:domain-generalization} shows that next-call
prediction transfers across domains, while end-task success is less stable. For
example, the SearchQA-trained Qwen3-4B checkpoint improves Airline/Retail Hit@1
from 35.6/39.7 to 45.6/50.1 after RL, while task success drops from 21.6/36.4
to 17.6/29.8. In the reverse direction, the ToolScale-trained Qwen3.5-4B
checkpoint improves HotpotQA/MuSiQue Hit@1 from 25.4/28.3 to 35.4/37.9 after
RL, while task success also decreases. This suggests that the training procedure
teaches transferable call-level behavior, such as matching common tool-call
arguments, but successful task completion still depends on domain-specific
reasoning and interaction strategy. Matched-domain agentic RL training is
therefore important for preserving task success in the target domain.

\begin{table}[t]
\centering
\renewcommand{\arraystretch}{1.08}
\small
\setlength{\tabcolsep}{1.2pt}
\resizebox{0.95\columnwidth}{!}{%
\begin{tabular}{@{}l|cc|cc|cc|cc@{}}
\toprule \midrule
\multirow{2}{*}{Model} &
\multicolumn{2}{c|}{\textbf{HotpotQA}} &
\multicolumn{2}{c|}{\textbf{MuSiQue}} &
\multicolumn{2}{c|}{\textbf{Tau-Air}} &
\multicolumn{2}{c}{\textbf{Tau-Ret}} \\
\cmidrule(lr){2-3} \cmidrule(lr){4-5} \cmidrule(lr){6-7} \cmidrule(lr){8-9}
 & H@1 & Succ. & H@1 & Succ. & H@1 & Succ. & H@1 & Succ. \\
\midrule
\rowcolor{transgray}
\multicolumn{9}{c}{\textbf{Qwen3-4B}} \\
\midrule
Base & 23.2 & \textbf{52.1} & 26.4 & \textbf{15.1} & 35.6 & \textbf{21.6} & 39.7 & \textbf{36.4} \\
SFT & 37.8 & 49.4 & 40.9 & 13.8 & 41.2 & 18.9 & 45.8 & 32.0 \\
RL & \textbf{55.2} & 48.6 & \textbf{58.4} & 12.9 & \textbf{45.6} & 17.6 & \textbf{50.1} & 29.8 \\
\midrule
\rowcolor{transgray}
\multicolumn{9}{c}{\textbf{Qwen3.5-4B}} \\
\midrule
Base & 25.4 & \textbf{63.1} & 28.3 & \textbf{25.1} & 42.1 & \textbf{56.4} & 46.8 & \textbf{69.7} \\
SFT & 31.8 & 60.0 & 34.2 & 22.8 & 58.6 & 53.1 & 62.9 & 66.0 \\
RL & \textbf{35.4} & 57.9 & \textbf{37.9} & 21.6 & \textbf{76.1} & 51.4 & \textbf{79.4} & 63.7 \\
\midrule
\bottomrule
\end{tabular}
}
\caption{Generalization across domains. We evaluate whether a self-speculator
trained in one interaction family transfers to the other.}
\label{tab:domain-generalization}
\vspace{-0.12in}
\end{table}

\subsection{Ablation Study}
\label{sec:ablation}

\begin{figure}[t]
\centering
\includegraphics[width=0.98\linewidth]{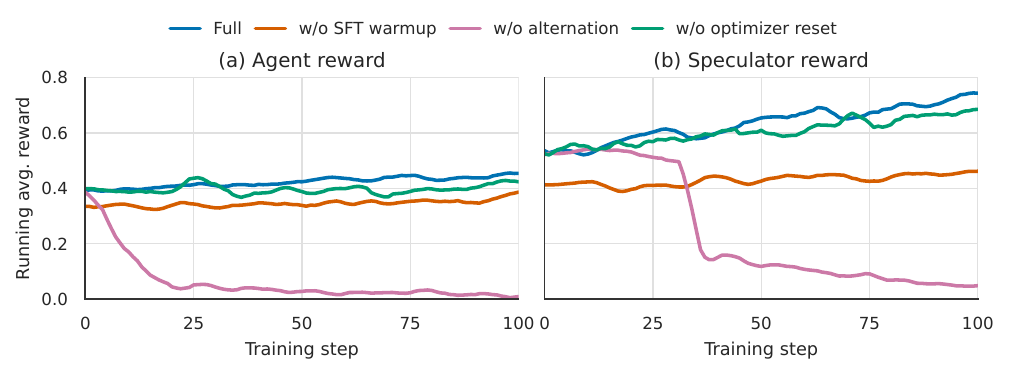}
\caption{Reward trend over the first 100 steps for each mode:
(left) agent reward and (right) speculator reward. The full method
preserves agent reward while improving speculator reward. Removing SFT warmup,
optimizer resets, or alternating updates leads to weaker or less stable reward trends.}
\label{fig:reward-ablation-trends}
\vspace{-0.14in}
\end{figure}

We ablate the stabilization techniques in the joint training recipe on SearchQA
with Qwen3-4B.
Figure~\ref{fig:reward-ablation-trends} shows training-time reward traces when
removing SFT warmup, optimizer reset, or alternating schedule, while
Table~\ref{tab:alternating-schedule} compares different agent:speculator update
schedules under the same iteration budget.

\paragraph{Effect of SFT warmup.}
The w/o SFT warmup variant starts RL directly from the base model, testing
whether the speculation suffix alone is enough to induce well-formed
call-prediction behavior. In Figure~\ref{fig:reward-ablation-trends}, this
variant lags the full method in speculation reward and is less stable early in
training. This suggests that warmup provides a useful initialization before RL
optimizes next-call prediction accuracy, because early updates can otherwise
spend capacity on discovering the output format rather than matching the
agent's eventual call.

\paragraph{Effect of optimizer reset.}
The w/o optimizer reset variant keeps the alternating schedule but preserves
optimizer state across mode switches. In
Figure~\ref{fig:reward-ablation-trends}, it produces weaker and less stable
reward trends than the full method. The two modes optimize different targets:
agent updates improve task-solving behavior, while speculator updates improve
next-call prediction. If the same optimizer state is carried from one mode into
the other, momentum and adaptive statistics from the previous target can push the
new updates in a less useful direction~\citep{sutskever2013momentum,kingma2015adam}.
Resetting the optimizer at each mode switch gives each block a fresh optimizer
state and makes training more stable by limiting cross-mode interference through
optimizer history.

\paragraph{Effect of alternating updates.}
The w/o alternating updates trace in Figure~\ref{fig:reward-ablation-trends}
shows that mixing agent and speculator updates leads to early collapse in training. Table~\ref{tab:alternating-schedule} compares block schedules under
the same iteration budget. Switching every step is worst: the 1:1 schedule
achieves only 31.8 average Hit@1 and 10.3 average task success. Performance
improves as each mode is trained for longer consecutive updates, reaching 55.2
average Hit@1 and 26.1 average task success with the 4:8 schedule. This suggests
that the speculator needs enough consecutive signal to track the agent's current
call distribution, while the agent still needs periodic updates to preserve task
performance.

\begin{table}[t]
\centering
\scriptsize
\setlength{\tabcolsep}{1.4pt}
\resizebox{0.95\columnwidth}{!}{%
\begin{tabular}{@{}r|cc|cc|cc|cc@{}}
\toprule \midrule
\multirow{2}{*}{\begin{tabular}{@{}c@{}}$m(k)$\\schedule\end{tabular}} &
\multicolumn{2}{c|}{\textbf{Hotpot}} &
\multicolumn{2}{c|}{\textbf{MuSiQ}} &
\multicolumn{2}{c|}{\textbf{BCP}} &
\multicolumn{2}{c}{\textbf{Avg.}} \\
\cmidrule(lr){2-3} \cmidrule(lr){4-5} \cmidrule(lr){6-7} \cmidrule(lr){8-9}
 & H@1$\uparrow$ & Succ.$\uparrow$
 & H@1$\uparrow$ & Succ.$\uparrow$
 & H@1$\uparrow$ & Succ.$\uparrow$
 & H@1$\uparrow$ & Succ.$\uparrow$ \\
\midrule
1:1 & 31.8 & 18.4 & 34.6 & 6.8 & 29.1 & 5.6 & 31.8 & 10.3 \\
2:2 & 41.9 & 39.6 & 44.7 & 13.9 & 38.8 & 15.2 & 41.8 & 22.9 \\
4:4 & 50.1 & 49.8 & 53.2 & 16.3 & 47.5 & 7.2 & 50.3 & 24.4 \\
4:8 & \textbf{55.2} & \textbf{53.0} & \textbf{58.4} & \textbf{17.1} & \textbf{52.1} & \textbf{8.1} & \textbf{55.2} & \textbf{26.1} \\
\midrule
\bottomrule
\end{tabular}
}
\caption{Ablation over agent:speculator update schedules on SearchQA. We report
Hit@1 and task success on the three search datasets used in the main table.}
\label{tab:alternating-schedule}
\vspace{-0.12in}
\end{table}

\section{Related Work}
\label{sec:relatedwork}

\paragraph{Reinforcement learning for agentic agents.}
Reinforcement learning is widely used to post-train LLMs, from RLHF and
AI-feedback alignment~\citep{ouyang2022instructgpt,bai2022constitutionalai} to
recent reasoning-oriented RL methods and systems~\citep{shao2024deepseekmath,deepseekai2025r1,kimiteam2025kimi,yu2025dapo},
including efficient reasoning, verifier-based code RL, and external tool use~\citep{hou2025thinkprune,liu2025harnessllm,he2026hardtestgen,schick2023toolformer,patil2023gorilla,chen2023fireact,liu2024apigen,liu2026agentskills}.
More recent work uses RL to optimize multi-turn tool behavior, including search
and web agents~\citep{jin2025searchr1,song2025r1searcher,tan2025ragr1,qi2025webrl,zheng2025deepresearcher},
general tool-integrated reasoning and function calling~\citep{qian2025toolrl,zhang2025toolr1,li2025torl,feng2025retool,singh2025artist},
embodied multi-robot control~\citep{ji2025robotcontrol}, and long-horizon
interactive agent training~\citep{chen2025longhorizoninteractive,zeng2025turnlevelcredit,luo2025agentlightning}.
Our work also uses RL for tool-using agents, but optimizes a different
objective: the same policy must both solve the task and predict its own future
tool calls from partial trajectories.

\paragraph{Tool call speculation.}
Speculative decoding accelerates language-model inference by verifying drafts
from a faster model~\citep{leviathan2023speculativedecoding}. Tool-call
speculation applies the same idea at the action level: likely future tool calls
are issued early so tool latency can be hidden behind model generation. Recent
work explores this direction with external speculators, tool caches,
action-specific strategies, workflow-pattern reuse, and system-level
scheduling~\citep{ye2026speculativeactions,nichols2025speculativetoolcalls,zhong2026dualspec,sui2026paste,huang2025spagent,hooper2026speculativeinteraction}.
These methods show that action-level speculation can reduce latency, but they
rely on a separate draft model, historical traces, or action-specific heuristics.
In contrast, our self-speculating agent uses the deployed model itself as the
speculator, avoiding a second model or trace cache while keeping speculation
targets on-policy.

\section{Conclusion}

In this paper, we introduced self-speculating agents, where the deployed model
also predicts its own next tool call through a speculation suffix. This design
addresses the speculator--agent gap of external draft models while avoiding
separate model and KV-cache overhead. We train the dual-mode agent with online
joint agent-speculator RL using the agent's own rollouts as speculation targets.
Across agentic SearchQA and conversational tool-use benchmarks, self-speculation
improves next-call prediction while preserving downstream task success, showing
that tool-call speculation can be folded into the agent itself.

\section*{Limitations}

While our results show that a self-speculating agent can improve next-tool-call
prediction while preserving downstream task performance, the present study is an
initial step rather than a complete treatment of speculative tool use. Our
method and evaluation still have several limitations that should be considered
when interpreting the results.

Our method assumes that speculative tool calls can be issued without changing the
external environment. This is appropriate for read-only tools such as search,
retrieval, database lookup, and many information-gathering API calls, where an
incorrect speculation can simply be discarded. It is less suitable for tools
that mutate external state, such as placing an order, updating a database
record, sending a message, or triggering an irreversible workflow. In such
settings, speculative execution would require additional safeguards, such as
dry-run modes, transactional rollback, human confirmation, or restricting
speculation to the read-only prefix of a task. Since a large fraction of agentic
tool use is read-only or has a read-only planning stage, we expect speculation
to remain broadly useful, but the deployment policy must distinguish safe calls
from state-changing calls.

Our empirical evaluation is also limited in scope. We evaluate self-speculation
on two agentic task families: multi-hop search question answering and
conversational tool use. These settings cover different tool-call patterns,
but they do not exhaust the diversity of agent environments, especially code
execution, web navigation, long-running workflows, and multi-agent tool use.
Moreover, our training experiments are conducted with 4B-scale backbone models.
Larger models may exhibit different optimization dynamics, different
speculator-agent gaps, and different sensitivity to the stabilization mechanisms
introduced in this work. Extending the evaluation to more domains and model
scales is therefore an important direction for future work.

\bibliography{custom}

\clearpage
\appendix

\section{Implementation Details}
\label{sec:appendix:implementation-details}

During evaluation, speculation is run at each realized tool-call boundary of the
agent trajectory. The speculator receives the prefix before the tool call and
predicts a candidate call, but the environment executes only the agent's actual
call. This keeps the evaluated trajectory identical to the ordinary agent
trajectory and makes each speculation label the current agent's own next action.
During RL, we use the same prefix--next-call construction to derive speculation
examples from freshly sampled agent rollouts, as described in
Section~\ref{sec:method:dual-mode-rollout}.

\subsection{Training Hyperparameters}
\label{sec:appendix:training-hyperparameters}

Table~\ref{tab:training-hyperparameters} lists the default training settings
used for joint agent-speculator RL. Unless stated otherwise, ablations use the
same rollout budget, learning rate, schedule, and evaluation protocol as the
corresponding full run.

\begin{table}[h]
\centering
\setlength{\tabcolsep}{4pt}
\resizebox{\columnwidth}{!}{%
\begin{tabular}{@{}ll@{}}
\toprule
Setting & Value \\
\midrule
Backbone & Qwen3-4B, Qwen3.5-4B \\
RL method & DAPO-style GPO \\
$G_{\mathrm{ag}}$ & 8 \\
$G_{\mathrm{sp}}$ & 8 \\
Schedule & 4 agent / 8 speculator \\
Reset & Each mode switch \\
Learning rate & $3\times10^{-6}$ \\
Training length limit & 16K tokens (SearchQA); 32K tokens ($\tau$-bench) \\
RL budget & 200 steps per setting \\
Training hardware & 8 H100 GPUs (SearchQA); 8 H200 GPUs (ToolScale) \\
SFT data & Successful trajectories and speculation examples \\
SFT target & Agent trajectories; next calls under the speculation suffix \\
\bottomrule
\end{tabular}
}
\caption{Default training hyperparameters.}
\label{tab:training-hyperparameters}
\end{table}

\subsection{Prompt and Tool Schema Details}
\label{sec:appendix:prompt-and-tool-schema-details}

The fixed speculation suffix used in all experiments is:
\begin{quote}
\small\ttfamily
\textless think\textgreater{} Okay, let's see. The user provided what I need.
I'll look it up. The next step is to make the tool call.
\textless/think\textgreater{}
\end{quote}

For each speculation query, the suffix is appended after the intermediate
trajectory prefix and before the model decodes the candidate tool call. We use
the same suffix for off-the-shelf speculator evaluation, SFT warmup example
construction, joint agent-speculator RL, and final speculation evaluation. The
suffix does not name a benchmark, tool, argument key, or target answer.

We represent every tool call as a structured action $a=(n,\sigma)$, where $n$
is the tool name and $\sigma$ is a dictionary of named arguments. SearchQA
actions contain a search tool name and query string. $\tau$-bench-style actions
use the API name and schema-constrained argument dictionary supplied by the
environment. Calls that cannot be parsed into the expected structured format are
treated as invalid predictions and receive zero speculation reward.

For Hit@1 evaluation, a prediction is correct only if the parsed tool name and
the complete parsed argument dictionary exactly match the agent's eventual call.
For the shaped speculation reward used during RL, we first require the tool name
to match and then compute macro token-F1 over target argument keys, as defined
in Equation~\ref{eq:reward}. Missing predicted argument keys receive zero score,
and extra predicted keys do not compensate for missing target keys. This keeps
the training reward dense enough to provide partial credit for close arguments,
while the deployment reuse metric remains exact structured-call equality.

\section{Artifact Licenses}
\label{sec:appendix:artifact-licenses}

Table~\ref{tab:artifact-licenses} summarizes the licenses reported by the
upstream publishers for the external artifacts used in our experiments. We use
these artifacts for research evaluation and model training under their published
terms, and we retain the corresponding citations and license notices when
redistributing metadata or scripts.

\begin{table}[h]
\centering
\small
\setlength{\tabcolsep}{4pt}
\begin{tabular}{@{}p{0.62\columnwidth}p{0.30\columnwidth}@{}}
\toprule
Artifact & Reported license \\
\midrule
Qwen3-0.6B, Qwen3-1.7B, Qwen3-4B &
Apache-2.0 \\
Qwen3.5-0.8B, Qwen3.5-2B, Qwen3.5-4B &
Apache-2.0 \\
Qwen3-Embedding-0.6B &
Apache-2.0 \\
HotpotQA &
CC BY-SA 4.0 \\
MuSiQue &
CC BY 4.0 \\
BrowseComp-Plus &
MIT \\
$\tau$-bench / $\tau^2$-bench &
MIT \\
ToolScale &
Upstream dataset terms \\
\bottomrule
\end{tabular}
\caption{Licenses or access terms reported by the upstream model, dataset, or
benchmark publishers for artifacts used in the experiments.}
\label{tab:artifact-licenses}
\end{table}

\section{Data Statistics}
\label{sec:appendix:data-statistics}

Table~\ref{tab:data-statistics} reports the train, evaluation, and SFT warmup
counts used in our experiments. SFT counts are separated into agent-trajectory
rows and speculation rows; slash-separated counts denote Qwen3-4B / Qwen3.5-4B
when both warmup files are used. Following the subset-based evaluation practice
in R1-Searcher~\citep{song2025r1searcher}, SearchQA evaluation uses
500 examples per dataset sampled from the FlashRAG development splits, rather
than the full public development or test sets. We do not assume that these are
the identical examples released by R1-Searcher unless their instance IDs match
explicitly in metadata.

\begin{table}[h]
\centering
\small
\setlength{\tabcolsep}{4pt}
\resizebox{\columnwidth}{!}{%
\begin{tabular}{@{}lcccc@{}}
\toprule
Resource & Train & Eval & SFT agent & SFT spec \\
\midrule
FlashQA / SearchQA total & 283,338 & 2,500 subset & 1,658 / 1,351 & 1,658 / 1,351 \\
HotpotQA & -- & 500 subset & -- & -- \\
MuSiQue & -- & 500 subset & -- & -- \\
BrowseComp-Plus & 680 & 150 & -- & -- \\
ToolScale & 4,063 & -- & 0 & 5,000 \\
$\tau$-bench Airline & -- & 50 & -- & -- \\
$\tau$-bench Retail & -- & 114 & -- & -- \\
\bottomrule
\end{tabular}
}
\caption{Dataset, benchmark, and SFT warmup counts. The ToolScale warmup source
contains speculation prompt--label rows only.}
\label{tab:data-statistics}
\end{table}

\section{LLM Usage Statement}
\label{sec:appendix:llm-usage}

We used large language models to assist with paper grammar editing and with code
implementation and debugging. 

\end{document}